\documentclass[letterpaper, 10 pt, journal, twoside]
{IEEEtran}
\IEEEoverridecommandlockouts      
\usepackage{floatpag}

\usepackage{cite}
\usepackage{float}
\usepackage{arydshln}
\usepackage{makecell}
\usepackage{balance}
\usepackage{graphics} 
\usepackage{epsfig} 
\usepackage{mathptmx} 
\usepackage{times} 
\usepackage{amsmath} 
\usepackage{amssymb}  
\usepackage{xcolor}
\usepackage{colortbl}
\usepackage[breaklinks,colorlinks]{hyperref}
\usepackage{caption}
\usepackage{subcaption}
\usepackage{multirow}
\let\labelindent\relax
\usepackage{enumitem}
\usepackage{booktabs}
\usepackage{physics}
\usepackage[normalem]{ulem}

\definecolor{mygreen}{rgb}{0.0, 0.5, 0.0}
\usepackage{tabularx, booktabs}
\usepackage{cleveref}
\usepackage{fancyvrb}

\author{
Yasaman Haghighi, Suryansh Kumar, Jean-Philippe Thiran, Luc Van Gool
\thanks{Suryansh Kumar and Luc Van Gool are with VIS Group ETH Z\"urich.}
\thanks{Yasaman  Haghighi and Jean Philippe Thiran are with EPFL Lausanne.}
}

\begin{document}

\title{Neural Implicit Dense Semantic SLAM}

\maketitle


\begin{abstract}
Visual Simultaneous Localization and Mapping (vSLAM) is a widely used technique in robotics and computer vision that enables a robot to create a map of an unfamiliar environment using a camera sensor while simultaneously tracking its position over time. In this paper, we propose a novel RGBD vSLAM algorithm that can learn a memory-efficient, dense 3D geometry, and semantic segmentation of an indoor scene in an online manner. Our pipeline combines classical 3D vision-based tracking and loop closing with neural fields-based mapping. The mapping network learns the SDF of the scene as well as RGB, depth, and semantic maps of any novel view using only a set of keyframes. Additionally, we extend our pipeline to large scenes by using multiple local mapping networks. Extensive experiments on well-known benchmark datasets confirm that our approach provides robust tracking, mapping, and semantic labeling even with noisy, sparse, or no input depth. Overall, our proposed algorithm can greatly enhance scene perception and assist with a range of robot control problems.

\end{abstract}



\section{Introduction}\label{sec:introduction}
The margin of this draft is limited to detail on the benefits of Simultaneous Localization and Mapping (SLAM) solutions to robotics \cite{durrant2006simultaneous}, computer vision \cite{cadena2016past}, control \cite{nuchter20076d}, and decision-making automation systems \cite{bala2022advances}. Even though SLAM could be solved using several types of sensing modalities \cite{khairuddin2015review}, in this paper, we focus on visual sensor measurements, i.e., RGB-D and RGB, to solve SLAM (popularly known as the V-SLAM problem), for an indoor scene. The goal of this problem is that given the visual input feed of the scene, the robot must be able to accurately localize its position and simultaneously build the scene's 3D map. Nevertheless, a V-SLAM approach that can provide scene geometry, camera position, and 3D semantic label simultaneously at inference time could greatly benefit robot perception and related downstream robot tasks \cite{menini2021real}. Classical vSLAM algorithms are able to estimate camera poses with high levels of accuracy and robustness in various scenarios while remaining fast and efficient in terms of computational costs. However, these algorithms build only a sparse point-based map of the environment. Having a dense map can help with scene understanding and allows the robot to interact more seamlessly with its surroundings. However, the significant amount of memory required to store such dense mappings makes them infeasible for practical use cases. Recently, Neural fields have demonstrated remarkable success in storing a continuous and dense representation of a scene using a limited amount of memory. As a result, there are new attempts to apply neural fields to SLAM \cite{nice-slam}\cite{eslam}. Such approaches optimize the map and camera poses by minimizing the rendering error of the network. However, these algorithms do not have loop closure to limit their tracking drift. Thus, they are not robust in challenging scenarios and do not extend to large scenes.

To overcome the above limitations and improve scene understanding, we propose a novel vSLAM algorithm based on ORB-SLAM 3 \cite{orb3} tracking and loop closing and combine it with a neural fields-based mapping. Our mapping network utilizes the backbone of Instant-NGP \cite{instant-ngp} due to its real-time performance, and we have further modified it based on NeuS \cite{neus} to enable learning the SDF of the environment for an accurate representation of the geometry. Our pipeline robustly tracks camera poses, provides accurate and dense geometry and semantic segmentation, and works in large scenes. Our main contributions are summarized below:
\begin{itemize}
    \item We propose an online algorithm to optimize the mapping network using a dynamic set of keyframes. The keyframes are selected based on \cite{orb3} criteria and constitute only a small portion of the entire dataset. This results in a significant speed improvement of the method and a reduction of its computational complexity.
    \item We introduce a novel approach for dense 3D semantic segmentation, based on 2D semantic color maps of keyframes. With this approach, we are able to accurately learn the dense 3D semantics of the scene online while simultaneously learning geometry. Our approach is robust, even when keyframe semantics are inconsistent and jittery.
    \item Our mapping network provides a multifaceted representation of SDF, semantics, RGB, and depth images while being memory efficient. For a room size of $25 \, \mathrm{m}^2$, we store all the information using less than 25 MB of memory.
    \item We extend our pipeline to large scenes by dividing the global space into subspaces and optimizing a separate map for the local coordinate system of each subspace.
    \item Finally, we conduct a comprehensive ablation study to confirm the robustness of our algorithm when dealing with noisy or sparse input depth, as well as when only RGB modality is available.
\end{itemize}

\begin{figure*}[t]
    \centering
\includegraphics[width=\textwidth]{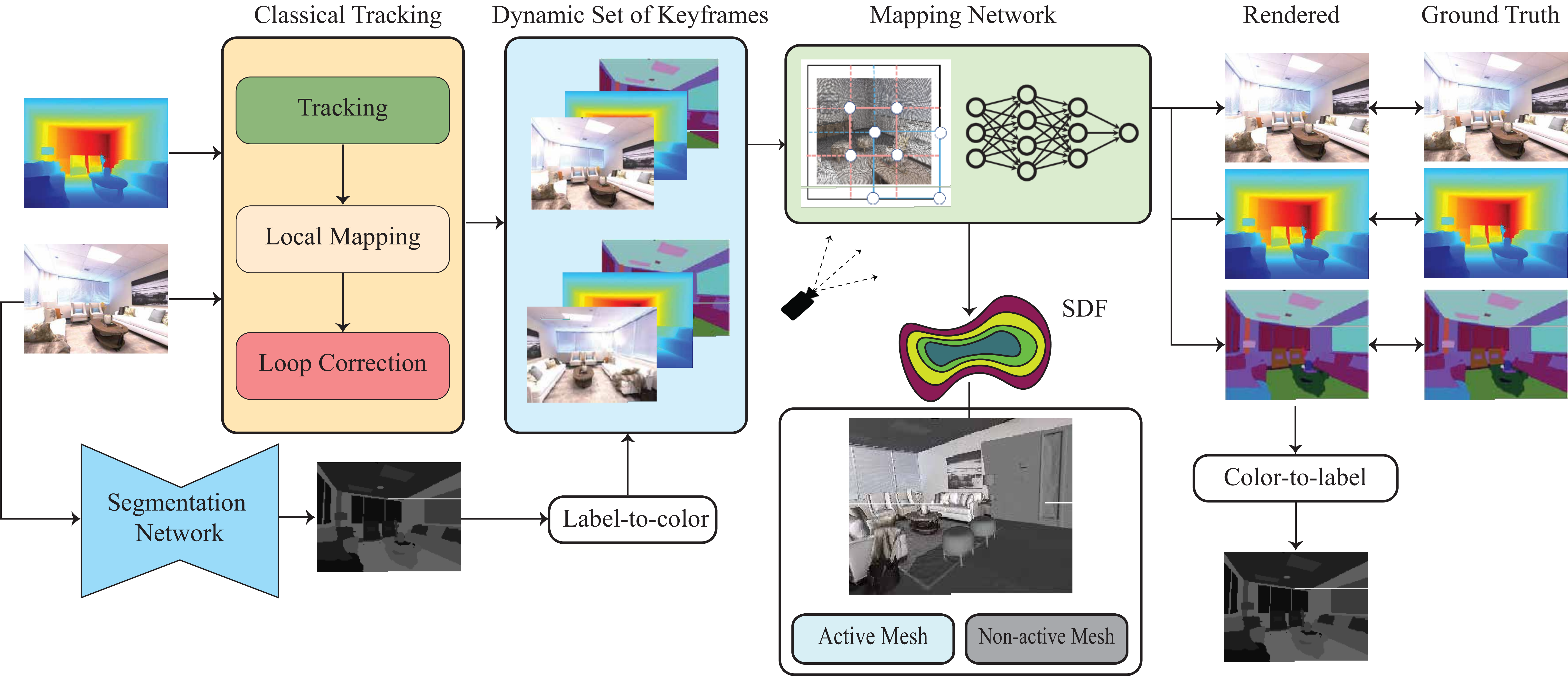}
\vspace{0.1cm}
    \caption{\small Overview of the RGBD-Semantic Pipeline: We begin by feeding each new frame into a classical tracking block, which estimates its camera pose and determines whether it should be considered a keyframe. A dynamic set of RGB, depth, and semantic segmentation of keyframes is maintained, with their poses being updated dynamically by the classical tracking block. To optimize the network, a keyframe is randomly selected from the dynamic set, and photometric, geometric, and semantic losses are used to optimize the network. Additionally, an active/non-active mesh paradigm is employed to handle situations where tracking is lost.}
    \label{fig:RGBD-semantic-pipeline-overview}
\end{figure*}

\section{Related Work}
The first monocular visual SLAM algorithm, MonoSLAM \cite{monoslam}, was a significant milestone in the field of visual SLAM, as it demonstrated the feasibility of using a single camera to build a map of an unknown environment and estimate the camera's position within that environment. Later, The Parallel Tracking and Mapping (PTAM) algorithm \cite{PTAM} separated the Tracking and Mapping components into two different threads, which allowed the algorithm to run in real time on a single processor. Moreover, PTAM introduced the concept of using keyframes for mapping, which allowed the algorithm to build a more accurate and robust map of the environment. Later, the large-scale direct monocular SLAM (LSD-SLAM) \cite{lsd-slam} was developed to address the  construction of large-scale maps. The ORB-SLAM \cite{orb3} algorithm is the state-of-the-art feature-based algorithm. It works in three parallel threads: tracking, local mapping, and loop closing. ORB-SLAM estimates accurate poses in different scenarios and extends to large scenes while being fast and efficient. However, it only builds a point-based map of the environment. In recent years, there has been a growing interest in building dense maps of the environment in SLAM using neural fields. iMAP \cite{imap} and NICE-SLAM \cite{nice-slam} use occupancy network to do both tracking and mapping and recently, ESLAM\cite{eslam} used a network to learn the TSDF of the scene. While the dense reconstruction of the above approaches is impressive, it is not yet real-time due to the network architecture they chose. Furthermore, their methods do not scale well to larger scenes. Lastly, optimizing both the map and camera pose using the same network may lead to getting stuck in local minima.

\section{Method}
 The overview of our pipeline is shown in \Cref{fig:RGBD-semantic-pipeline-overview}. In this section, we first describe our mapping network, which is followed by an explanation of our keyframe selection criteria. Next, we detail our online semantic segmentation approach. Finally, we present our end-to-end pipeline and discuss how it can be extended to handle larger-scale scenes.

\subsection{Mapping}
SLAM requires a mapping that can accurately represent the geometry of the environment while being memory-efficient and real-time. We use \cite{instant-ngp} as our backbone due to its real-time performance and memory efficiency. To improve the geometry representation, we modify the network by \cite{neus} to learn the SDF of the environment. Additionally, we decided not to feed the view direction to the color MLP, as it has been shown to decrease the accuracy of the geometry representation. Similar to \cite{neus}, we use volume rendering to compute color and depth values. Given the camera position $\mathbf{o}$ and a ray direction $\mathbf{v}$, we first sample $N$ points along the ray $\{\mathbf{p}(t_i) = \mathbf{o} + t_i\mathbf{v} | i=1,\ldots,N\}$ and accumulate color and depth values via 
\begin{equation}
\hat{I} = \sum_{i=1}^{n} T_i\alpha_i c_i \quad \mathrm{and} \quad \hat{d} = \sum_{i=1}^{n} T_i\alpha_i t_i,
\end{equation}
where
$
T_i = \prod_{j=1}^{i-1} (1 - \alpha_j)
$
is the accumulated transmittance and $\alpha_i$ is the opacity given by
\begin{equation}
\alpha_i = \max \left (\frac{\Phi_s(f(\mathbf{p}(t_{i}))) - \Phi_s(f(\mathbf{p}(t_{i+1})))}{\Phi_s(f(\mathbf{p}(t_i)))}, 0 \right).
\label{discrete_alpha}
\end{equation}
 Here, $f(.)$ is the SDF function and $\Phi_s(.)$ represents the sigmoid operation, i.e. $\Phi_s(x) = \operatorname{sigmoid}{(sx)}$. 

We optimize the network on a set of randomly selected $p$ pixels using a combination of photometric and geometric losses as given in the following equation:
\begin{equation}
    \begin{aligned}
L &= L_{\mathrm{photometric}} + L_{\mathrm{geometric}} \\
&= \sum_p \norm{I_{\mathrm{gt}}(p) - \hat{I}(p)}_1 + \sum_p \norm{d_{\mathrm{gt}}(p) - \hat{d}(p)}_2
\end{aligned}
\label{eq:rgbd-loss}
\end{equation}
where the sum is taken over the selected pixels, and the geometric loss is computed only for pixels with non-zero depth values.
Once the network is optimized, it provides a multifaceted representation of RGB, depth, and normal maps from any novel views. Additionally, in contrast to classical TSDF-based approaches that require a fixed voxel size, here, we can sample SDF with any desirable resolution to generate the mesh. 
\subsection{Keyframe selection}
When a room is scanned at a rate of 15 or 30 FPS, a large number of frames are often highly correlated with one another. Since neural fields are known for accurate view interpolation, we argue that optimizing the network on a set of keyframes is sufficient. To select keyframes, we rely on \cite{orb3} criteria. This approach has two advantages:
\begin{enumerate}
    \item Optimizing the network on a set of keyframes reduces the number of frames by approximately a factor of 10, resulting in a significant acceleration of the algorithm.
    \item ORB-SLAM performs pose graph optimization and bundle adjustment only for keyframes, and the poses of other frames are updated relative to keyframes. Therefore, optimizing the network on keyframes results in more accurate mapping.  
\end{enumerate}
Our ablation study demonstrates that the utilization of keyframes does not compromise the accuracy of the network in terms of geometry and semantic representation.
\subsection{3D semantic segmentaion}
Accurate semantic segmentation of a scene plays a crucial role in scene understanding and robot-scene interaction. However, achieving accurate 3D semantic segmentation is a challenging task that requires significant computational resources. On the other hand, 2D semantic segmentation is a well-studied problem and we have reliable networks that can segment anything accurately \cite{SAM}. We propose a new segmentation algorithm to learn the dense 3D semantics of the scene online using only the 2D semantics of keyframes. To accomplish this, we incorporate a segmentation decoder into our mapping network, whereby the texture and segmentation decoders utilize a common geometry block. Unlike similar approaches \cite{semanticnerf}\cite{panoptic} that use class probabilities for estimating semantic labels of a scene, we first convert the segmentation maps into color encodings and then use the encodings to optimize our neural field. The intuition behind our semantic segmentation approach is inspired by a simple example. Imagine a room with green flooring, blue walls, and red furniture. In such a scenario, we can easily segment the room by thresholding the colors. Since neural fields are highly expressive in learning the color of a 3D scene from multi-view data, using color maps for segmentation takes advantage of this strong prior and helps us predict robust segmentation maps even when the 2D semantics are jittery and inconsistent across keyframes. Also, as we rely only on color maps for each frame, we can easily extend the pipeline to panoptic segmentation by assigning different colors to each prediction instance. 

Our semantic segmentation pipeline consists of the following stages:
\begin{enumerate}
\item We use a 2D semantic segmentation network to retrieve the semantic segmentation of each keyframe.
\item The semantic segmentations are converted to colormaps.
\item We optimize our modified network architecture using photometric, geometric, and semantic losses:
\begin{equation}
   \resizebox{\linewidth}{!}{$L =  \sum_p \norm{I_{\mathrm{gt}}(p) - \hat{I}(p)}_1 + \sum_p \norm{d_{\mathrm{gt}}(p) - \hat{d}(p)}_2 + \sum_p \norm{s_{\mathrm{gt}}(p) - \hat{s}(p)}_2$}.
\end{equation}
Similar to \Cref{eq:rgbd-loss}, the depth loss is calculated only for pixels with non-zero depth values. Moreover, we assign black color to the pixels with unknown semantic labels and compute the semantic loss only for pixels with non-black color encodings.
\item After network optimization, we can render the semantic colormaps from any novel viewpoint and convert them to labels.
\end{enumerate}

\subsection{Online RGBD pipeline}
Our online RGBD SLAM pipeline, shown in \Cref{fig:RGBD-semantic-pipeline-overview}, consists of the following main stages:
\begin{itemize}
\item RGBD ORB-SLAM 3 \cite{orb3} for tracking and loop closing
\item A dynamic list of keyframes and their poses
\item Weighted random keyframe selection
\item Mapping network 
\item Active/non-active meshes
\end{itemize}
More specifically, we first feed each new frame into RGBD ORB-SLAM 3 to track the frame and determine whether it is a keyframe or not. We maintain a dynamic set of keyframes, with the keyframe list and their poses being dynamically updated by ORB-SLAM. To optimize the network, we randomly select a keyframe from the dynamic set of keyframes and optimize the network using photometric, geometric, and semantic losses on a set of $p$ pixels. Our random selection process is weighted, assigning higher weights to keyframes that have recently been added to the set of keyframes. Furthermore, we have an active/non-active mesh paradigm to be compatible with the ORB-SLAM multi-maps Atlas. Whenever ORB-SLAM tracking is lost and the algorithm starts a new map, we save the current network parameters and begin optimizing a new network.

\subsection{Extension to large scenes}
An effective SLAM algorithm should perform accurately in large scenes. The tracking capability of our pipeline extends to large scenes as ORB-SLAM is designed to handle them. However, the mapping network is the bottleneck in our pipeline. While simply increasing the size of the multi-resolutional hash encoding of \cite{instant-ngp} may seem like a straightforward solution to extend our pipeline to larger scenes, the network's expressivity will eventually impose a limit on the level of scene details it can represent. To overcome this limitation, given the ORB-SLAM global space $S$, we divide $S$ into mutually exclusive subspaces $S_i$ that can be predefined by the user at the beginning of the algorithm. We use a $5 \times 5 \times 5  \,
 \textrm{m}^3$ cube as the default subspace and represent each subspace by its center $c_i$. Each subspace has a dynamic set of keyframes and a specific mapping network. When ORB-SLAM detects a new keyframe, we use back-projection to find the view frustum of the frame and assign it to the corresponding subspace. If the keyframe is lying on the overlap of more than one set, we add it to all of the sets. For optimizing the mapping network of each subspace, given a camera position $t_G$ in the global coordinate system, we transform $t_G$ from the global coordinate to the subspace coordinate system via $t_{S_i} = t_G - c_i$ and work with the local coordinate system of $S_i$. To generate meshes during inference, we first compute the sub-mesh for each subspace $S_i$ and then transform the mesh to the global coordinate system.

\begin{figure}[t!]
    \centering
    \includegraphics[width=0.48\textwidth]
    {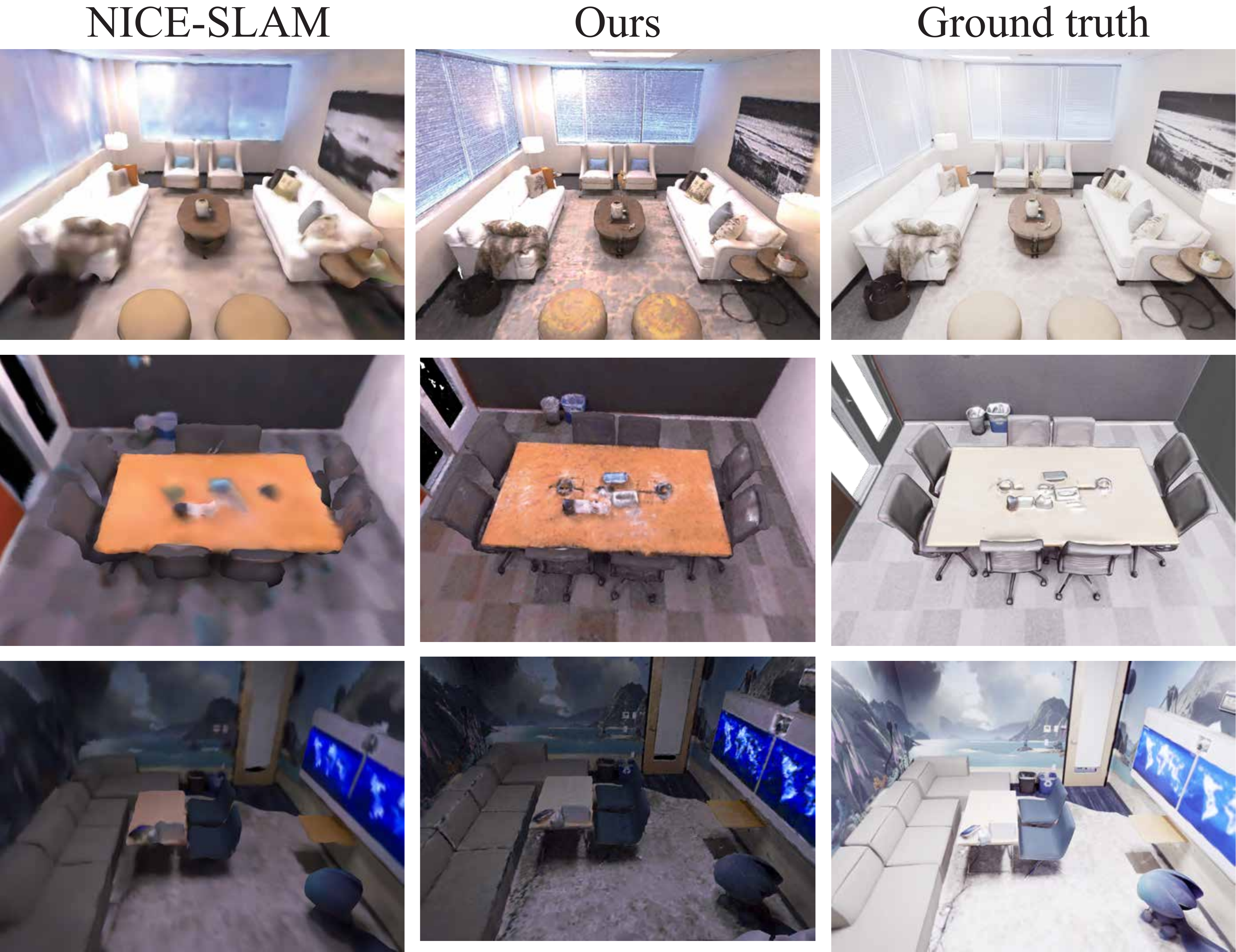}
    \caption{\small Qualitative evaluation of 3D reconstruction on Replica dataset.} \label{rendered-mesh}
\end{figure}

\begin{table}[t]
    \centering
    \resizebox{0.48\textwidth}{!}{
    \begin{tabular}{cccccccccc}
    \toprule
      & \tt{rm-0}  & \tt{rm-1} & \tt{rm-2} & \tt{off-0} & \tt{off-1} & \tt{off-2} & \tt{off-3} & \tt{off-4} & Avg. \\
    \midrule
    iMAP \cite{imap} & 5.23 & 3.09 & 2.58 & 2.40 & 1.17 &  5.67 & 5.08 & 2.23 & 3.42 \\
    NICE-SLAM \cite{nice-slam} &  1.69 & 2.13 & 1.87 & 1.26 & 0.84 & 1.71 & 3.98 & 2.82 & 2.05\\
    ESLAM \cite{eslam} & 0.71  & 0.70 & \textbf{0.52} & \textbf{0.57} & 0.55 &  \textbf{0.58} &  \textbf{0.72} & \textbf{0.63} &  \textbf{0.63}\\
    Ours & \textbf{0.58} & \textbf{0.41} & 0.58 & 0.62 & \textbf{0.40 }& 1.20 & 0.88 & 1.80 & 0.80
    \\
    \bottomrule
    \end{tabular}
    }
    \caption{\small Tracking evaluation on Replica RGB-D dataset. We report ATE RMSE [cm] for each room and different methods. The table shows that ORB-SLAM 3 \cite{orb3} can provide accurate and robust tracking for all rooms. The values of other methods are taken from \cite{eslam}.} \label{table:ate-rgbd-orb}
\end{table}

\begin{figure}[t]
    \centering
    \includegraphics[width=0.48\textwidth]
    {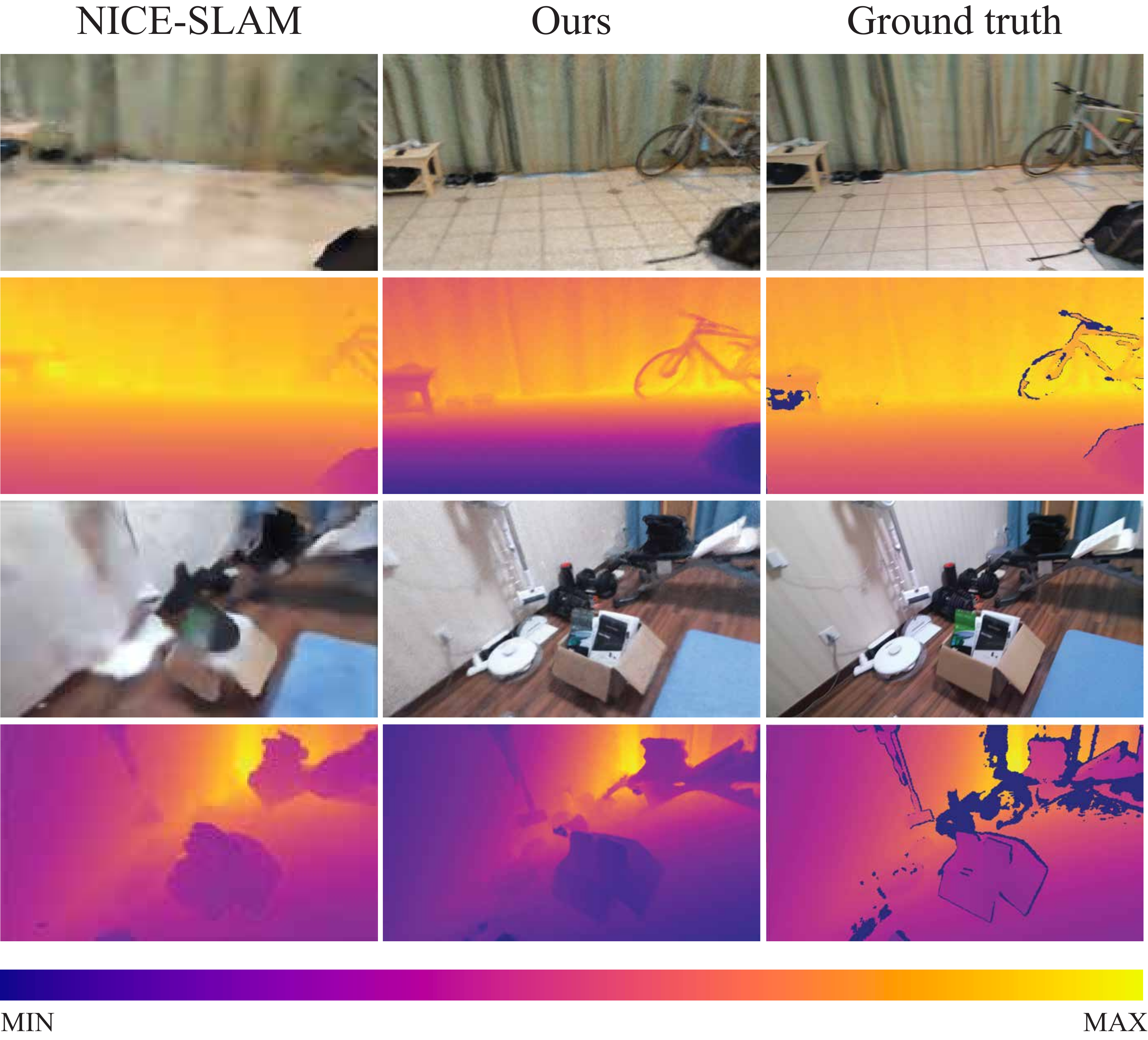}
    \caption{\small A qualitative evaluation of our rendered RGB and depth images for ScanNet and Apartment datasets.} \label{rendered-mesh-scannet2}
\end{figure}

\begin{table}[t]
    \centering
    \resizebox{0.5\textwidth}{!}{
    \begin{tabular}{cccccccccc}
    \toprule
      & \tt{rm-0} & \tt{rm-1}  & \tt{rm-2}  & \tt{off-0} & \tt{off-1} & \tt{off-2} & \tt{off-3} & \tt{off-4}& Avg.\\
    \midrule
    TSDF Fusion &6.38 & 5.33 & 6.84 & 4.74 & 4.62 & 11.32 & 9.89 & 6.49 & 6.95   \\
     iMAP \cite{imap}& 5.70 & 4.93 & 6.94 & 6.43 & 7.41 & 14.23 & 8.68 & 6.80 & 7.64\\
     NICE-SLAM \cite{nice-slam}& 2.11 & 1.68 & 2.90 & 1.83 & 2.46 & 8.92 & 5.93 & 2.38 & 3.53 \\
     DI-Fusion \cite{DI-fusion}& 6.66 &  96.82 &  36.09 &  7.36 &  5.05 &  13.73 &  11.41 &  9.55 &  23.33 \\
     ESLAM \cite{eslam} & 0.97 &  1.07 &  1.28  &  0.86 &  1.26 &  1.71 &  1.43 &  \textbf{1.06} &  1.18 \\
     Orbeez-SLAM \cite{orbeez} &  - &  - &  - &  - &  - &  - &  - &  - &  11.88 \\
     Ours &   \textbf{0.51} &  \textbf{0.32} &  \textbf{0.44} &  \textbf{0.33} &  \textbf{0.34} &  \textbf{0.53} &  \textbf{0.69} &  1.39 &  \textbf{0.56} \\
    \bottomrule
    \end{tabular}
    }
    \caption{\small Geometry reconstruction evaluation for the Replica dataset. We compute the L1 error [cm] between the estimated and ground truth depth maps. On average, our method outperforms other algorithms by at least by a factor of two. Voxel Resolution used for TSDF is 512. The values of other methods are taken from \cite{nice-slam, eslam}.} \label{table:replica-depth-eval}
\end{table}

\begin{table}[tp]
  \renewcommand{\arraystretch}{1.4}   
  \centering%
    \resizebox{0.48\textwidth}{!}{
    \begin{tabular}{llcccccccccccccccccc}
    \toprule
    & &  \multicolumn{1}{c}{\makecell{\tt{rm-0}}} & \multicolumn{1}{c}{\makecell{\tt{rm-1}}} &  \multicolumn{1}{c}{\makecell{\tt{rm-2}}} & \multicolumn{1}{c}{\makecell{\tt{off-0}}} & \multicolumn{1}{c}{\makecell{\tt{off-1}}} & \multicolumn{1}{c}{\makecell{\tt{off-2}}}& \multicolumn{1}{c}{\makecell{\tt{off-3}}} & \multicolumn{1}{c}{\makecell{\tt{off-4}}} & Avg. \\
    \midrule    
    \multirow{4}{*}{\rotatebox[origin=c]{90}{\scriptsize NICE-SLAM \cite{nice-slam}}}
    & PSNR [dB]$\uparrow$ & 23.83 & 22.61 & 21.97 & 25.78 & 25.30 & 18.50 & 22.82 & 25.26 & 23.26\\
    & SSIM$\uparrow$ & 0.788 & 0.813 & 0.858 & 0.887 & 0.842 & 0.826 & 0.862 & 0.875 & 0.844 \\
    & LPIPS$\downarrow$ & 0.284 & 0.249 & 0.218 & 0.209 & 0.145 & 0.242 & 0.190 & 0.191 & 0.216 \\
    & & & & & & & & & &  \\
    \midrule
    \multirow{4}{*}{\rotatebox[origin=c]{90}{\scriptsize Vox-Fusion \cite{yang2022vox}}}
    & PSNR [dB]$\uparrow$ & 23.45 & 20.83 & 18.38 & 23.28 & 24.48 & 17.50 & 23.06 & 24.84 & 21.98\\
    & SSIM$\uparrow$ & 0.765 & 0.773 & 0.747 & 0.751 & 0.762 & 0.727 & 0.824 & 0.851 & 0.775 \\
    & LPIPS$\downarrow$ & 0.280 & 0.272 & 0.282 & 0.235 & 0.169 & 0.292 & 0.232 & 0.212 & 0.247 \\
    & & & & & & & & & &  \\
    \midrule
    \multirow{4}{*}{\rotatebox[origin=c]{90}{\scriptsize COLMAP \cite{colmap}}}
    & PSNR [dB]$\uparrow$ & 20.93 & 11.67 & 10.35 & 5.88 & 5.88 & 15.66 & 13.73 & 17.47 & 12.70\\
    & SSIM$\uparrow$ & 0.778 & 0.698 & 0.757 & 0.595 & 0.539 & 0.806 & 0.792 & 0.811 & 0.722 \\
    & LPIPS$\downarrow$ & 0.291 & 0.443 & 0.330 & 0.444 & 0.337 & 0.303 & 0.299 & 0.273 & 0.340 \\
    & & & & & & & & & &  \\
    \midrule
     \multirow{4}{*}{\rotatebox[origin=c]{90}{\scriptsize Droid-SLAM \cite{droid}}}
    & PSNR [dB]$\uparrow$ & 18.25 & 18.65 & 13.49 & 16.13 & 10.31 & 14.78 & 15.53 & 15.71 & 15.36\\
    & SSIM$\uparrow$ & 0.737 & 0.793 & 0.786 & 0.760 & 0.650 & 0.800 & 0.797 & 0.800 & 0.765 \\
    & LPIPS$\downarrow$ & 0.352 & 0.283 & 0.299 & 0.298 & 0.286 & 0.300 & 0.302 & 0.311 & 0.304 \\
    & & & & & & & & & &  \\
    \midrule
    \multirow{4}{*}{\rotatebox[origin=c]{90}{\scriptsize Nicer-SLAM \cite{nicer-slam}}}
    & PSNR [dB]$\uparrow$ & 25.64 & 23.69 & 22.62 & 25.88 & 22.56 & 21.46 & 24.42 & 25.15 & 23.93\\
    & SSIM$\uparrow$ & 0.810 & 0.820 & 0.871 & 0.885 & 0.828 & 0.863 & 0.888 & 0.887 & 0.857 \\
    & LPIPS$\downarrow$ & 0.254 & 0.233 & 0.200 & 0.193 & 0.160 & 0.203 & 0.175 & 0.192 & 0.201 \\
    & & & & & & & & & &  \\
    \midrule
    \multirow{4}{*}{\rotatebox[origin=c]{90}{Ours}}
    & PSNR [dB]$\uparrow$ &  \textbf{33.16} & \textbf{35.18} & \textbf{36.49} & \textbf{40.22} & \textbf{38.90} & \textbf{34.22} & \textbf{34.74}&\textbf{33.24} & \textbf{35.76} \\
    & SSIM$\uparrow$ &  \textbf{0.987} & \textbf{0.987} &\textbf{0.991} &\textbf{0.993} &\textbf{0.973} & \textbf{0.933}& \textbf{0.994}&\textbf{0.988} & \textbf{0.980}\\
    & LPIPS$\downarrow$ &  \textbf{0.016}& \textbf{0.012} & \textbf{0.011}& \textbf{0.007}& \textbf{0.018}&\textbf{0.009} &\textbf{0.006} &\textbf{0.015} &\textbf{0.011}\\
    & & & & & & & & & &  \\
    \bottomrule
  \end{tabular}
  }
  \caption{\small Perceptual quality of the RGB images rendered from different methods. We can see that our method significantly outperforms the current state-of-the-art models in terms of visual fidelity. The values for other methods are taken from \cite{nicer-slam}.} \label{table:replica_rec_RGB}
\end{table}

\section{Experiments and Results}
We provide a comprehensive evaluation of our pipeline on Replica \cite{replica} and ScanNet \cite{scannet} datasets. To assess our SLAM performance for large scenes, we leverage captures of an apartment with multiple rooms provided by \cite{nice-slam}. Furthermore, we conduct extensive ablation studies to demonstrate the robustness of our pipeline for sparse depth images and noisy semantics.

\textbf{Evaluation Metric.}
For camera tracking, we employ evo toolbox \cite{evo} to first align the camera poses with ground truth poses and then compute the ATE RMSE \cite{sturm2012benchmark}. To evaluate the quality of our mapping, we report the L1 depth error [cm] and assess the PSNR, SSIM \cite{SSIM}, and LPIPS \cite{LPIPS} for rendered color images. To evaluate our semantic segmentation quality, we report accuracy, mIoU, and fwIoU \cite{IoU}.

\textbf{Implementation details}
The mapping network uses a multi-resolution hash-encoding with a hash table size of $2^{19}$. This is followed by a small vanilla MLP with one hidden layer to get the final signed distance function (SDF). The output of the SDF branch is then fed into separate fully fused MLPs from \texttt{tiny-cuda-nn} \cite{tiny-cuda-nn} for obtaining the color and segmentation values. As segmentation is easier to learn compared to RGB images, the segmentation branch employs two hidden layers, while the color branch has three hidden layers. All networks are optimized using ADAM optimizer \cite{adam} with a learning rate of $10^{-2}$, linear warm-up for the first $2k$ iterations, and exponential decay after the warm-up steps.

\subsection{Tracking, Mapping and Rendering Evaluation}
The tracking performance of our pipeline on Replica dataset \cite{replica} is summarized in \Cref{table:ate-rgbd-orb}. ORB-SLAM 3 \cite{orb3} tracks camera poses accurately, while remaining real-time and using minimal computational resources. Furthermore, our ablation study shows that ORB-SLAM 3 retains its robustness even when presented with noisy or sparse input depth data, as well as in large scenes.
The evaluation of our mapping reconstruction and rendering quality are summarized in \Cref{table:replica-depth-eval} and \Cref{table:replica_rec_RGB} respectively. In contrast to previous works, our network is optimized only on keyframes that are 10 times less than the full dataset. Our mapping generalizes well on unseen views and outperforms previous approaches. Additionally, a qualitative evaluation of our meshes is shown in \Cref{rendered-mesh}. \Cref{rendered-mesh-scannet2} also shows an example of our rendered RGB and depth images for the ScanNet dataset. Compared with NICE-SLAM \cite{nice-slam}, our mapping learns a  more detailed texture and geometry.

\subsection{Semantics Evaluation}
\begin{figure}[tp]
    \centering
    \includegraphics[width=0.48\textwidth]{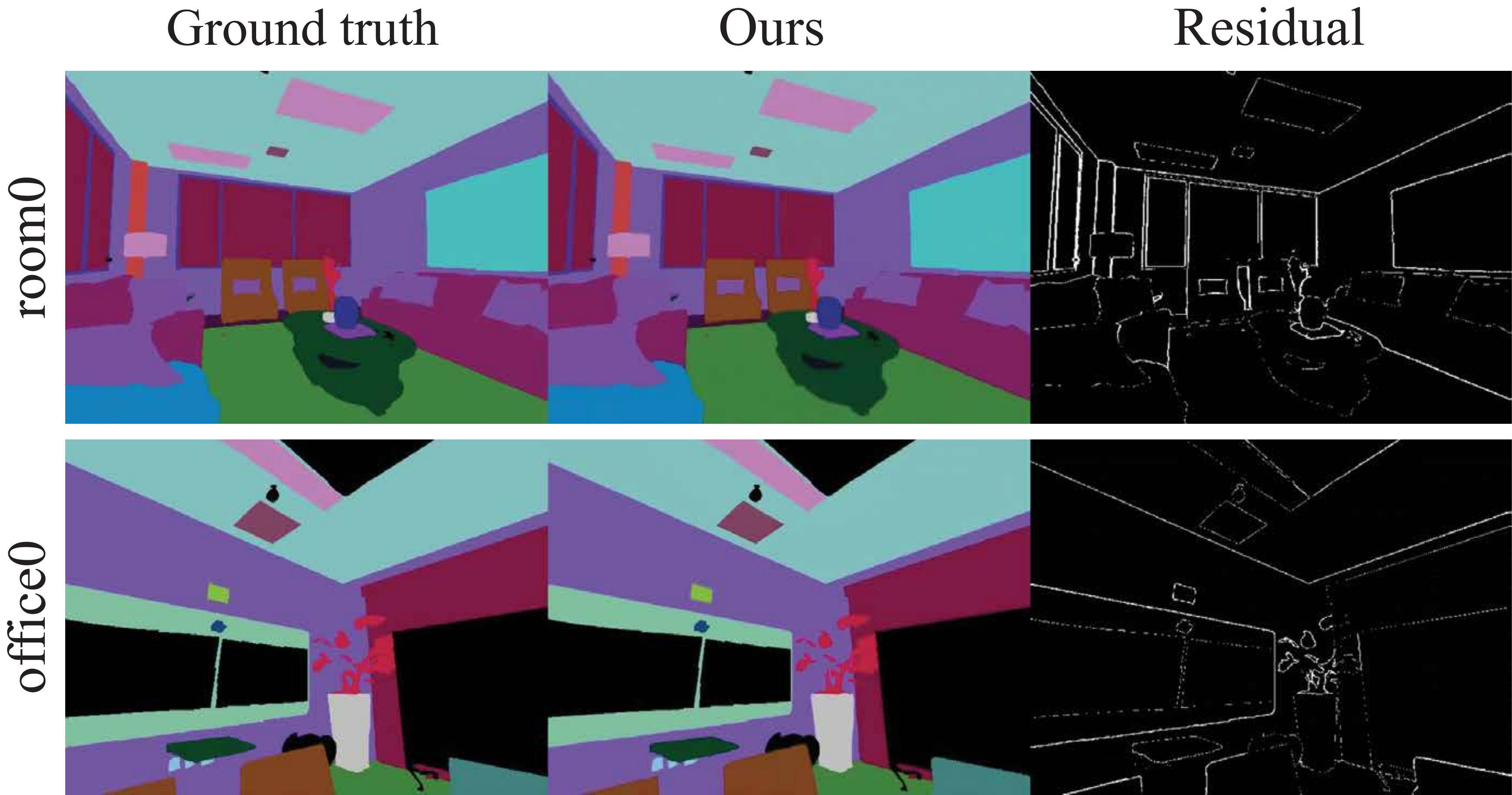}
    \caption{\small Semantic segmentation qualitative evaluation.}\label{fig:segments_qualitative_eval}
\end{figure}

\begin{figure}[tp]
    \centering
    \includegraphics[width=0.48\textwidth]{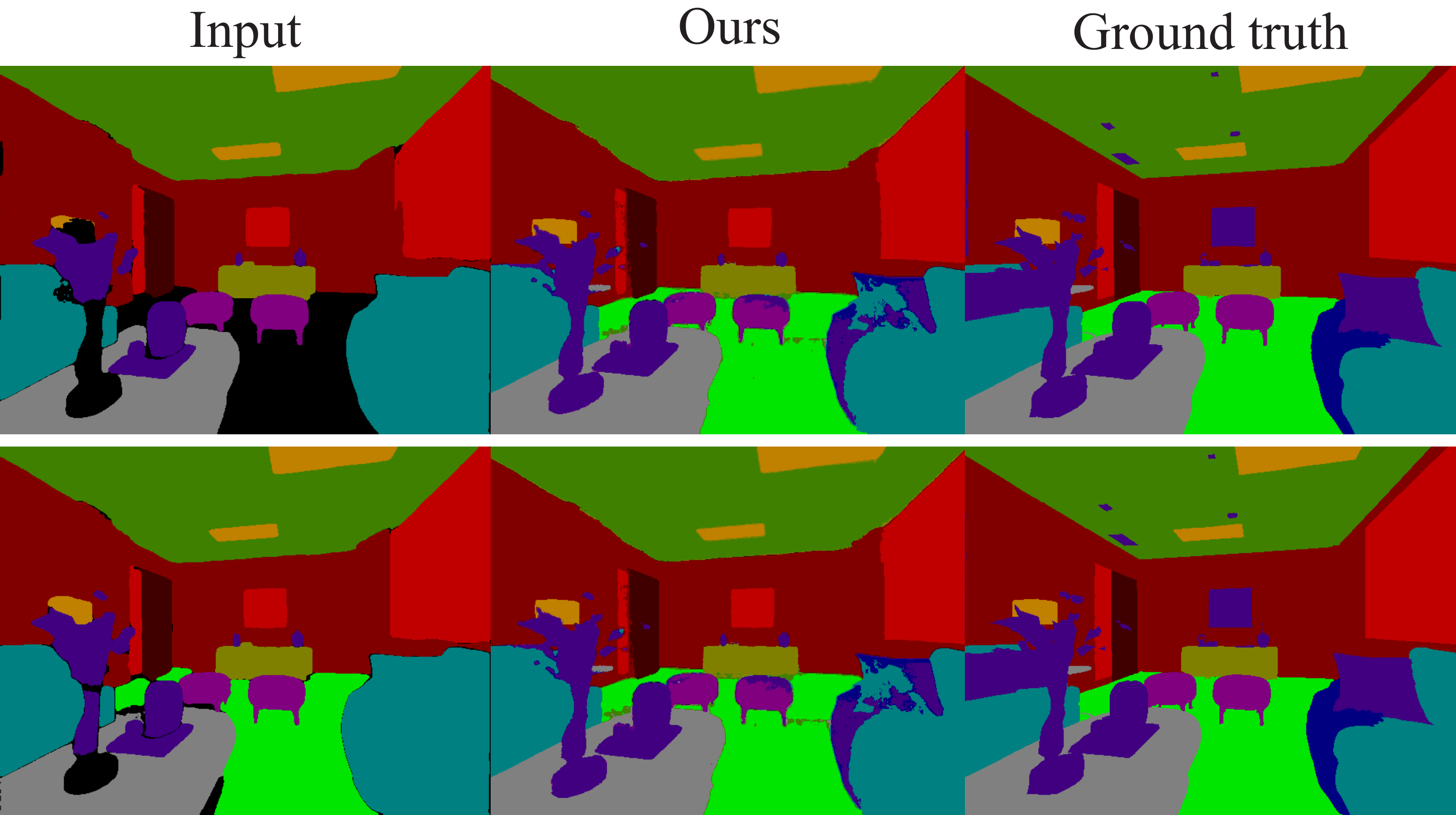}
    \caption{\small Semantic segmentation evaluation with inconsistent input semantics. Our approach results in consistent semantic segmentation, e.g. for the floor, and avoid assigning jittery labels to different areas.}\label{fig:seg-inconsis}
    \label{semantics2}
\end{figure}

\begin{figure}[t]
    \centering
    \includegraphics[width=0.5\textwidth]{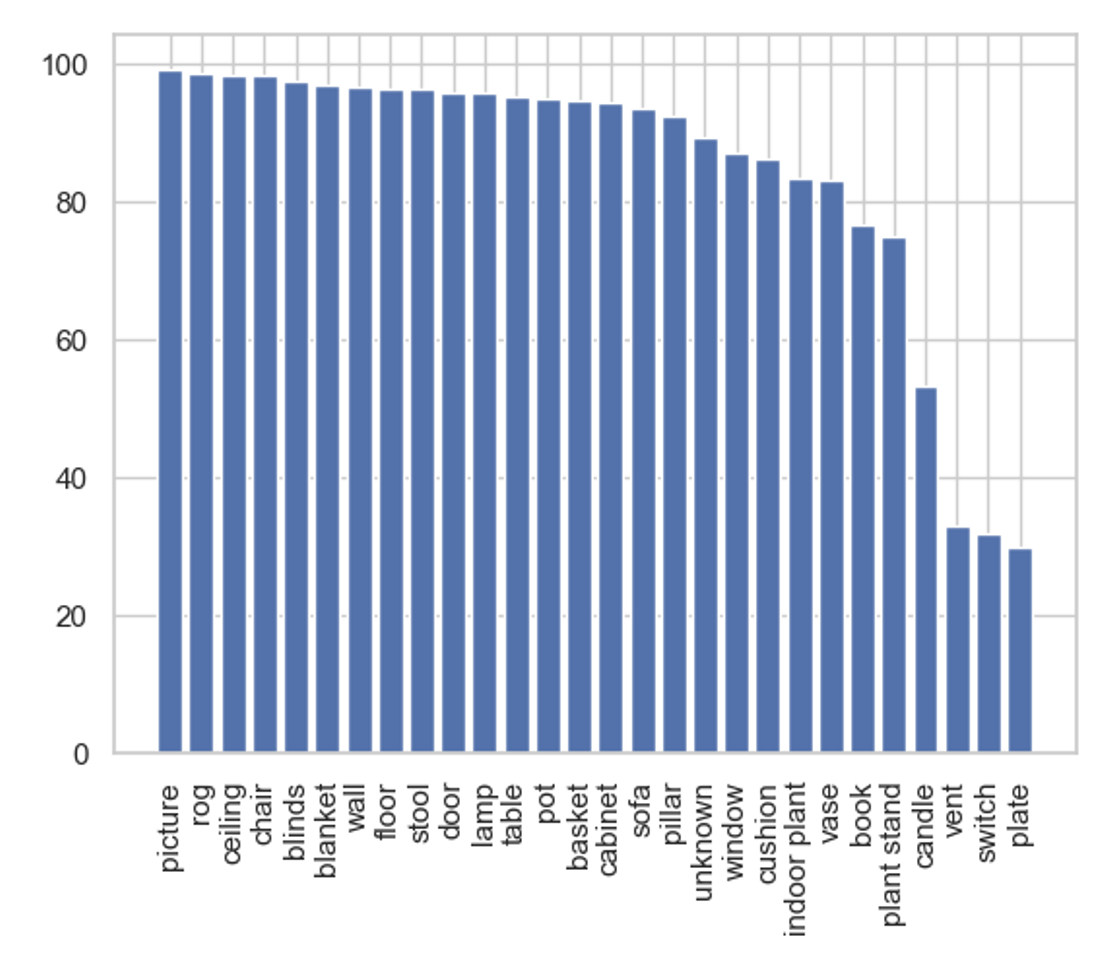}
    \caption{\small Per-class IoU percentage for room0. We can see that the semantic network only makes noticeable errors on tiny objects in the scene.}.
    \label{per-class-iou}
\end{figure}

To assess the quality of our semantic segmentation, we utilize the ground truth semantics of the keyframes to optimize our network. The quantitative and qualitative evaluation of our approach shown in \Cref{table:semantics-eval} and \Cref{fig:segments_qualitative_eval} respectively prove that we are able to accurately segment the 3D space with errors occurring only at the boundaries of objects. The per-class IoU of room 0 is shown in \Cref{per-class-iou} confirming that the error is high only for small classes. An important question arises: What happens to our pipeline if we use a 2D semantic segmentation network to estimate the keyframe semantics and these semantics are inconsistent? To address this question, we utilize \cite{m2f} to estimate keyframe semantics as discussed in \cite{panoptic}. We observe that our approach can estimate consistent semantics using only 2D inconsistent semantics of keyframes without utilizing additional losses or regularizers. The quantitative and qualitative evaluation of our approach are shown in \Cref{table:replica-sem-inconsistant} and \Cref{fig:seg-inconsis} respectively. For more visualizations, please refer to our supplementary video. Also, \Cref{table:semantics-vs-RGBD} demonstrates that the accuracy of the network's geometry representation is not affected by noisy semantics since we use a geometric loss during training. Hence, our segmentation method can fix the semantic inconsistencies across different views while maintaining an accurate geometric representation of the scene. 
\begin{table}[ht]
    \centering
    \caption{\small Semantic segmentation evaluation on Replica dataset.}
    \begin{tabular}{cccccc}
    \toprule
      & \tt{rm-0} & \tt{rm-1}  & \tt{rm-2}  & \tt{off-0} \\
    \midrule
    \tt{\# Keyframes} & 171 & 211 & 186 & 73   \\
     \tt{Total Accuracy (\%)}& 97.76 & 98.50 & 98.76 & 98.89 \\
     \tt{Class Avg. Accuracy (\%)}& 93.66 & 94.95 & 91.63 & 96.32\\
     \tt{mIOU (\%)}& 82.45 & 84.08 & 76.99 & 85.94 \\
     \tt{fwIOU (\%)} & 95.85 & 97.21 & 97.82 & 98.01\\
    \bottomrule
    \end{tabular}
    \label{table:semantics-eval}
\end{table}

\begin{table}[tp]
    \centering
    \begin{tabular}{cc}
    \toprule
      & mIoU \\
    \midrule
     Mask2Former \cite{mask2former} & 52.4\\
     SemanticNeRF \cite{semanticnerf} & 58.5\\
     DM-NeRF \cite{dm-nerf} & 56.0\\
     PNF \cite{PNF} & 51.5 \\
     Panoptic Lifting \cite{panoptic} & 67.2\\
     Ours & \textbf{67.3}\\
    \bottomrule
    \end{tabular}
    \caption{\small The mIoU on Replica dataset for inconsistent semantic input. The value of other methods are taken from \cite{panoptic}.}\label{table:replica-sem-inconsistant}
\end{table}

\begin{table}[t]
    \centering
    \resizebox{0.48\textwidth}{!}{
    \begin{tabular}{lcccccc}
    \toprule
      & & \tt{\# frames} & \tt{TA(\%)}  & \tt{CAA (\%)}  & \tt{mIoU(\%)}& \tt{fwIoU(\%)} \\
    \midrule
    \multirow{2}{*}{\tt{rm-0}} 
    
     &  key & 171 & 97.76 & 93.66 & 82.45 & 95.85  \\
     & full & 900 & 98.30 & 95.88 & 85.69 & 96.84   \\
     \midrule
     \multirow{2}{*}{\tt{rm-1}}
     & key & 211 & 98.50 & 94.95 & 84.08 & 97.21 \\
     & full & 900 & 99.34 & 96.61 & 87.03 & 98.81 \\
     \midrule
     \multirow{2}{*}{\tt{rm-2}}
     & key & 186 & 98.76 & 91.63 & 76.99 & 97.82\\
     & full & 900 & 99.25 & 93.39 & 80.17 & 98.74\\
     \midrule
     \multirow{2}{*}{\tt{off-0}}
     & key & 73 & 98.89 & 96.32 & 85.94 & 98.01 \\
     &  full & 900 & 99.15 & 97.21 & 87.50 & 98.50 \\
    \bottomrule
    \end{tabular}
    }
    \caption{\small Semantic segmentation evaluation on Replica dataset when optimizing the network on the full dataset and using keyframes only. The table shows that keyframes are enough for estimating accurate semantic segmentation maps. \texttt{TA} is total accuracy and \texttt{CAA} is class average accuracy.}\label{table:semantics-eval-full-vs-key}
\end{table}

\subsection{Robustness to depth map sparsity}
In this section, we investigate the impact of depth sparsity on our pipeline. Since each depth camera has its unique depth noise model, demonstrating the robustness of our pipeline to depth sparsity can allow us to handle uncertain depth pixels by setting their values to zero, thus preventing error propagation. To evaluate the effect of depth sparsity, we randomly select a percentage of pixels in each depth image and set their depth values to zero. We then use these sparse depth images as input to our pipeline.

\begin{figure}[t]
    \centering
    \includegraphics[width=0.35\textwidth]{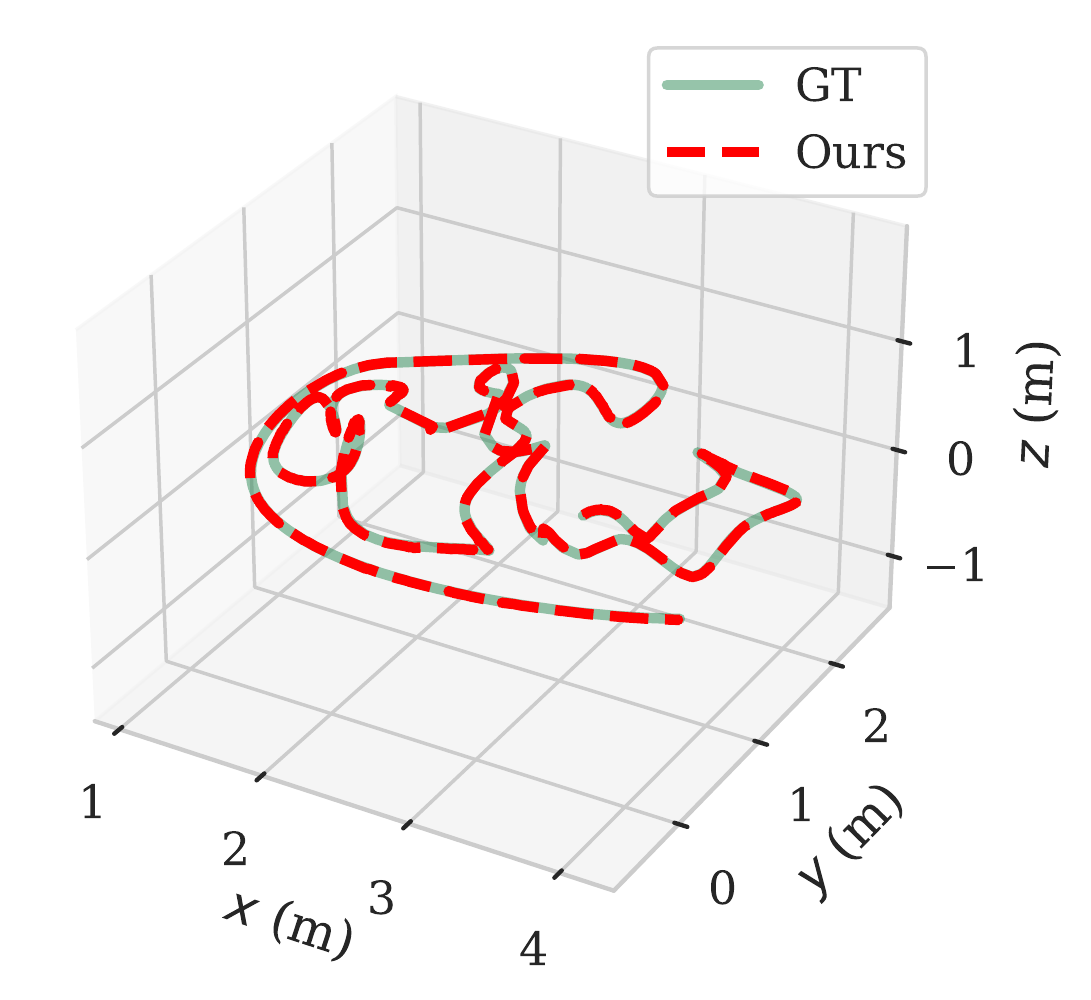}
    \caption{\small The effect of 70\% depth image sparsity on ORB-SLAM keyframe tracking for room0. The ORB-SLAM poses are highly accurate even when the input data is extremely sparse.}
    \label{70-depth-sparsity-poses}
\end{figure}

\begin{figure}[t]
    \centering
        \includegraphics[width=0.48\textwidth]{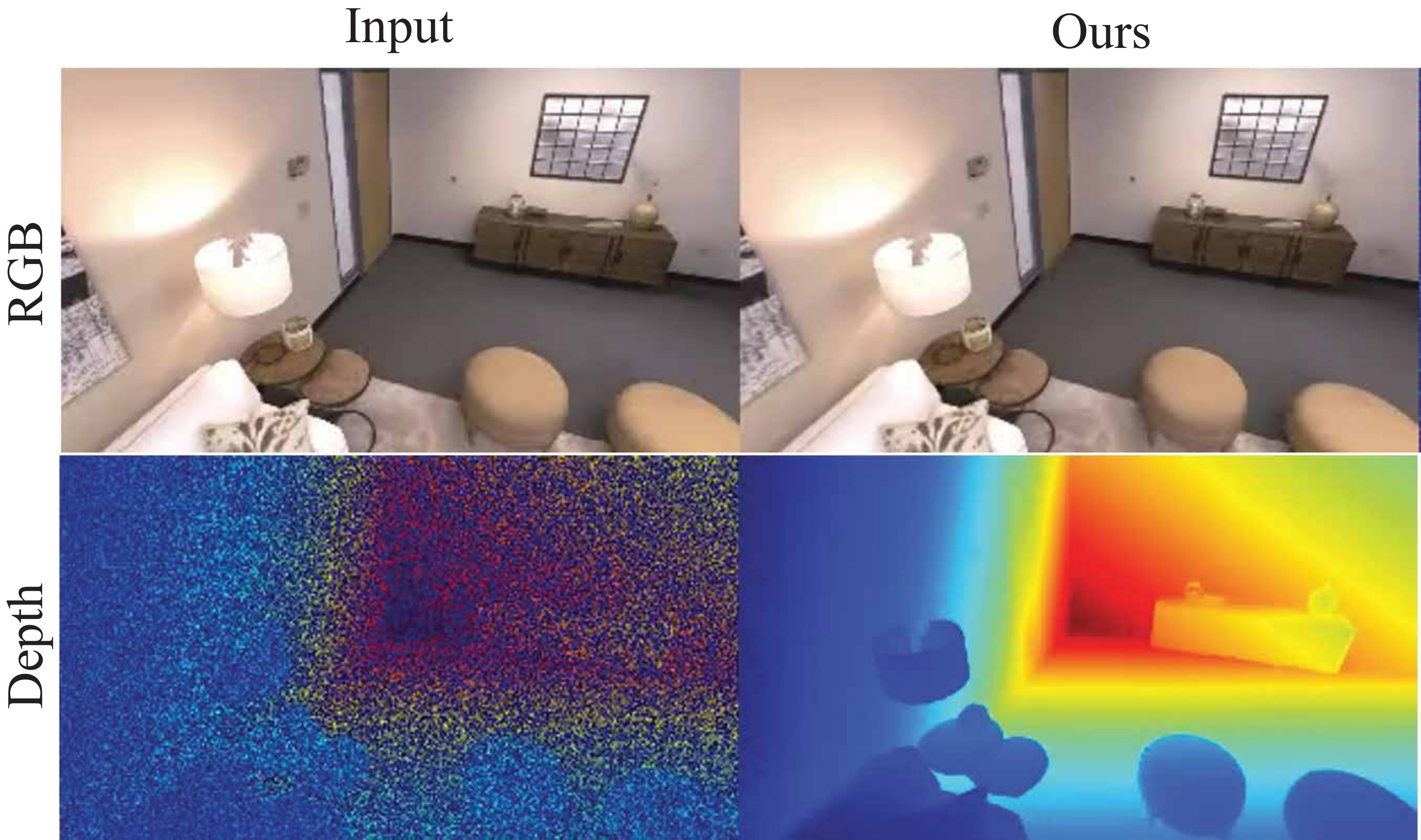}
    
    \caption{\small The effect of 70\% depth image sparsity on the mapping network. The network is able to perform high-quality view synthesis and depth estimation even when the input depth maps are sparse.}\label{fig:sparsity}
\end{figure}

The ATE RMSE and Mean for different sparsities are reported in \Cref{table:depth-sparsity-poses}. In addition, \Cref{70-depth-sparsity-poses} provides a qualitative comparison between ORB-SLAM keyframe poses and ground truth poses when 70\% sparsity is present. Our experiments prove that ORB-SLAM tracking is robust to depth sparsity. We then proceed to evaluate the impact of depth sparsity on our mapping. Our experiments reveal that as sparsity increases, ORB-SLAM increases the number of keyframes, which in turn results in optimizing our mapping network using more frames. Interestingly, the L1 depth error remains nearly constant for different sparsity percentages. These results confirm the robustness of our approach to depth sparsity. We summarize our quantitative findings in \Cref{table:depth-sparsity-mapping}. The \Cref{fig:sparsity} demonstrate the impact of 70\% depth sparsity on network-rendered RGB and depth images, revealing that our mapping network can accurately inpaint the missing depth information while generating accurate RGB images. 
\begin{table}[t]
    \centering
    \begin{tabular}{ccc}
    \toprule
      \tt{Sparsity} & \tt{ATE RMSE(cm)} & \tt{ATE Mean(cm)}\\
    \midrule
    \tt{0\%} & 0.55 & 0.52 \\
     \tt{10\%}& 0.57 & 0.54\\
     \tt{30\%}& 0.50 & 0.48 \\
     \tt{50\%}& 0.71 & 0.68\\
     \tt{70\%} & 0.65 & 0.60\\

    \bottomrule
    \end{tabular}
    \caption{\small The effect of depth image sparsity on ORB-SLAM keyframe tracking for room0. The table proves that ORB-SLAM is robust to depth sparsity and can perform accurate localization with highly sparse inputs.}
    \label{table:depth-sparsity-poses}
\end{table}

 \begin{table}[t]
    \centering
    \begin{tabular}{ccc}
    \toprule
      \tt{Sparsity} & \tt{\# keyframes} & \tt{L1 depth error (cm)}\\
    \midrule
    \tt{0\%} & 200 & 0.51 \\
     \tt{10\%}& 223 & 0.52\\
     \tt{30\%}& 285 & 0.51 \\
     \tt{50\%}& 572 & 0.58\\
     \tt{70\%} & 876 & 0.56\\
    \bottomrule
    \end{tabular}
    \caption{\small The effect of depth image sparsity on mapping network for room0. The L1 depth error remains almost constant even when the input depth maps are highly sparse.}
    \label{table:depth-sparsity-mapping}
\end{table}

\subsection{Extension to RGB modality}
\begin{figure}[htp]
    \centering
    \includegraphics[width=0.48\textwidth]{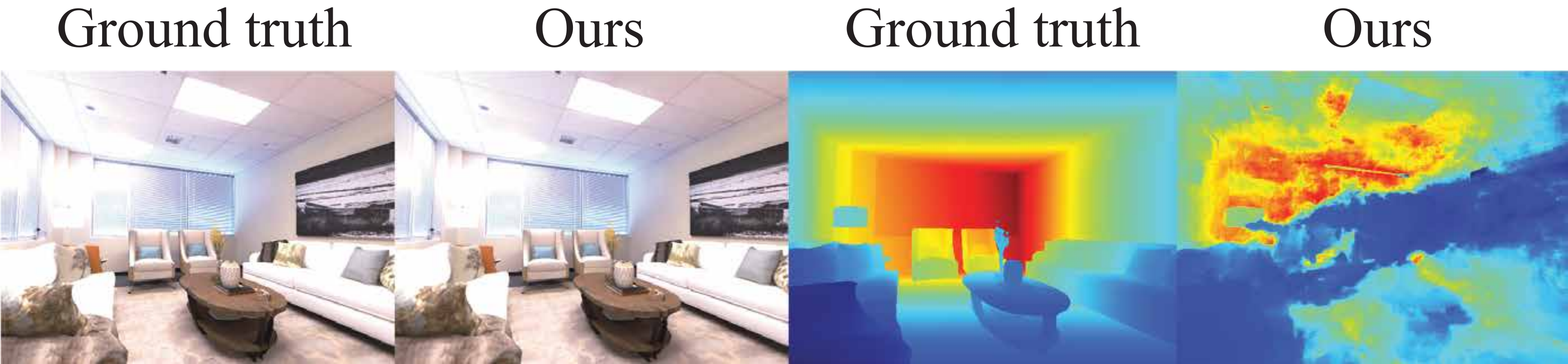}
    \caption{When using keyframes for RGB setup, the network is able to learn the RGB information accurately. However, it struggles to learn the proper geometry.}
    \label{fig:RGB-keyframe}
\end{figure}

\begin{figure}[htp]
    \centering
    \includegraphics[width=0.48\textwidth]{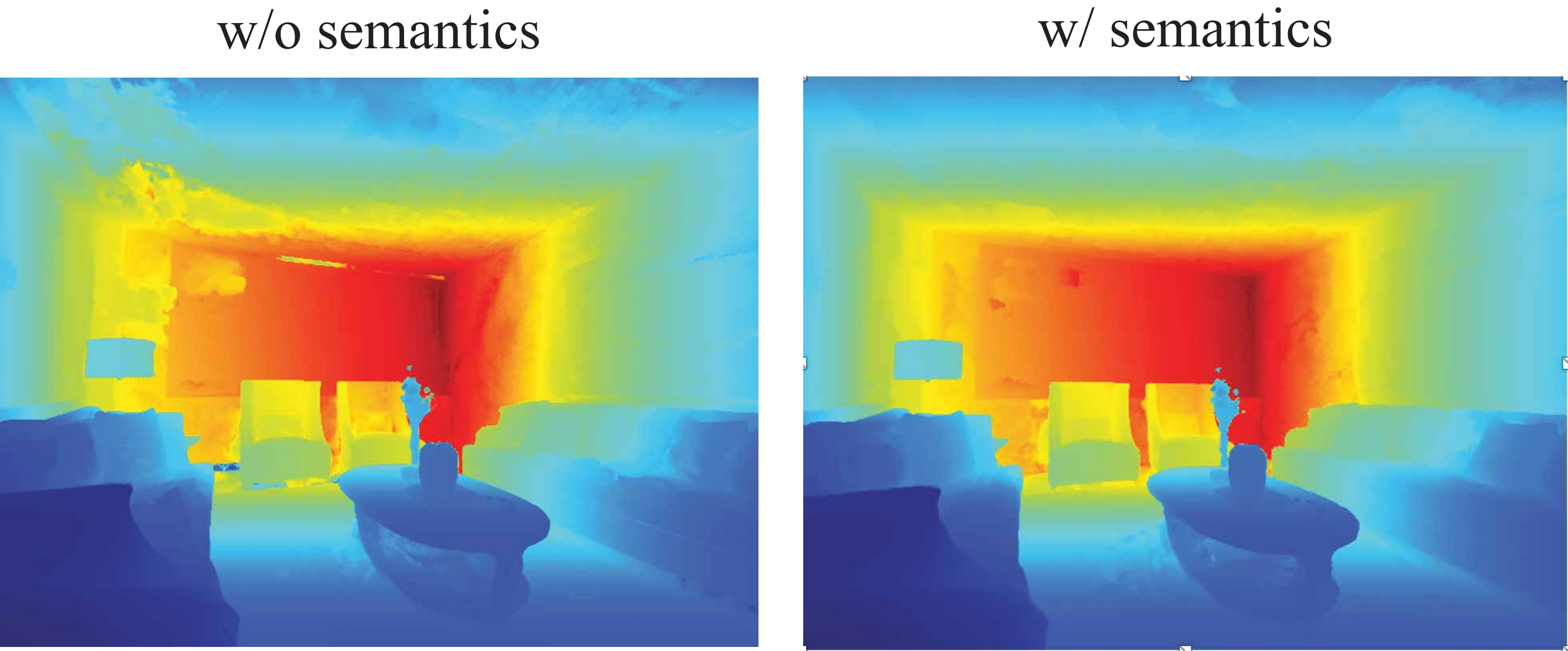}
    \caption{The semantic loss can successfully improve the learned geometry, especially for flat regions of the scene.}\label{RGBS}
\end{figure}

To expand our pipeline to the RGB modality, a straightforward approach would be to use RGB ORB-SLAM to retrieve camera poses up to a scale ambiguity and optimize the network only with the photometric loss. However, the RGB pipeline presents the following challenges:
\begin{enumerate}
\item As shown in \Cref{fig:RGB-keyframe}, the pipeline can perform accurate novel view synthesis. However, the network is unable to estimate the correct geometry from keyframes alone. Thus, it requires all frames of the dataset to learn the geometry.
\item Even when using all frames, the network needs to be optimized for approximately five times more iterations than the RGBD setup to converge.
\item Additionally, each iteration takes longer in the RGB setup because, without depth information, we cannot accurately determine the empty space in a scene. As a result, the \cite{instant-ngp} rendering scheme that skips empty space doesn't work to its fullest potential.
\item Even when using the entire dataset to optimize the network with photometric loss, we observe that the network struggles to learn accurate geometry for flat surfaces like walls under varying indoor lighting conditions. An example of a depth image rendered by the network is shown in \Cref{RGBS}. 
\end{enumerate}
All of these factors make the algorithm slow and unsuitable for real-time applications.  One naive solution could be utilizing monocular depth estimation networks. However, if the network is not optimized on the same dataset, the estimated depth maps may not have geometric consistency, making it unrealistic for real-world scenarios where the SLAM algorithm is used for new scenes. To improve the learned geometry, we propose a new pipeline by combining semantic segmentation color maps with RGB information. As the semantic color representation and color images share the same geometry, the network learns the coarse scene representation from semantics and the fine details from RGB color. As a result, the network can better estimate the geometry of the scene. An example of a rendered depth image when including the semantic loss during training is given \Cref{RGBS}, and our quantitative results are summarized in \Cref{table:replica-depth-eval-seg}. Additionally, our early experiments revealed that in contrast to the RGBD pipeline, incorporating semantics into the RGB pipeline can only be beneficial if the semantic information is accurate. However, this assumption is reasonable given recent advancements in the field, as demonstrated by \cite{SAM}.

\subsection{Extension to large scenes}
\begin{figure}[htp]
    \centering
    \includegraphics[width=0.4\textwidth]{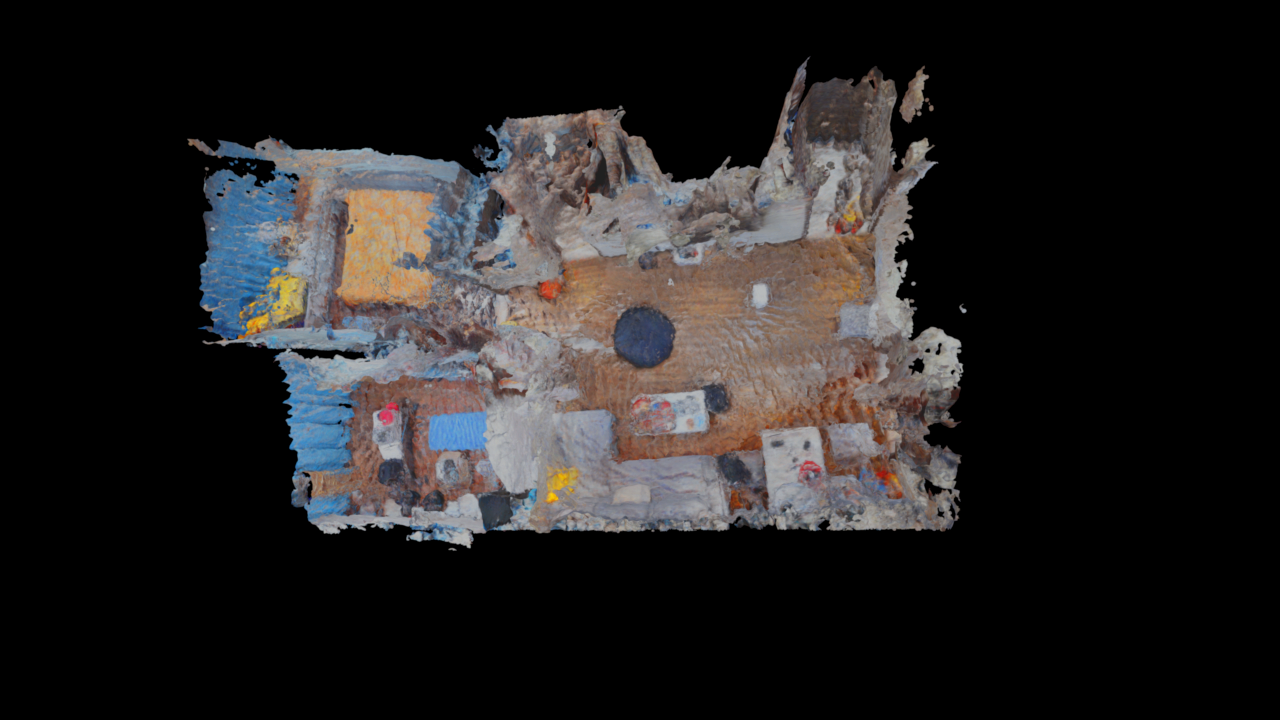}
    \caption{Reconstructed Apartment mesh.}
    \label{Apartmetn-mesh}
\end{figure}

To assess the performance of our pipeline on large scenes, we utilize the apartment dataset, which was captured by \cite{nice-slam} using an Azure Kinect camera. The color images in this dataset are blurry, and the depth images are both noisy and contain depth holes. While there are 19,595 images in the dataset, our keyframe list contains only 2,446 frames. We divide the apartment into three subspaces and optimize a separate network for each part. We compare our rendered RGB and depth images with NICE-SLAM in \Cref{rendered-mesh-scannet2} and show our reconstructed mesh in \Cref{Apartmetn-mesh}. The figure shows that our approach offers a more accurate and detailed color and geometry reconstruction.

\subsection{Runtime and memory usage analysis}
To evaluate the speed and size of our network, we follow \cite{eslam} and report the average frame processing time and the number of network parameters in \Cref{table:runtime}. For our pipeline, the average frame processing time varies during the optimization. It is because \cite{instant-ngp} rendering gets faster over time as it learns geometry. Approximately, it changes between 10 to 25 frames per second.

We observe that our network contains more parameters than previous works, but it is able to process frames faster. Moreover, our approach employs only keyframes, which constitute approximately 10 times less data than the full dataset. This results in even faster convergence.
\begin{table}[htp]
    \centering
    \begin{tabular}{ccc}
    \toprule
    & \tt{FPT(s)} & \tt{\#parameters (M)}\\
    \midrule
    iMAP & 5.20 & \textbf{0.22}  \\
     NICE-SLAM& 2.10 & 12.18  \\
     ESLAM & 0.18 & 6.79 \\
    Ours & \textbf{0.04 - 0.1} & 12.6  \\
    \bottomrule
    \end{tabular}
    \caption{\small Runtime and memory usage analysis. Our method is significantly faster than the other approaches thanks to the performance optimizations in the Instant-NGP blocks and ORB-SLAM real-time performance. The values of other methods are taken from \cite{eslam}.} \label{table:runtime}
\end{table}
\section{Ablations}
\subsection{Effect of using keyframes}
In this section, we investigate the impact of using keyframes instead of the full dataset on the quality of our geometry representation and semantic segmentation. To achieve this, we initially optimize the network using information from the entire dataset and then using only keyframes. We utilize ORB-SLAM pose estimates for both scenarios and evaluate the performance of the optimized networks on the same camera views. Our experiments summarized in \Cref{table:key-vs-full} and \Cref{table:semantics-eval-full-vs-key} demonstrate that we can achieve the same level of accuracy using only keyframes that are approximately 10 times less than the full dataset. 
\begin{table}[htp]
    \centering
    \begin{tabular}{cc||cc|cc}
    \toprule
      & \tt{key} & \tt{full}  & \tt{$\Delta$} & \tt{GT} & \tt{$\Delta$} \\
    \midrule
    \tt{rm-0} & 0.51 & 0.49 & 0.02  & 0.29 & 0.22 \\
     \tt{rm-1}& 0.32 & 0.27 & 0.05 & 0.19 & 0.13 \\
     \tt{rm-2}& 0.44 & 0.41 & 0.03 & 0.27 & 0.17 \\
     \tt{off-0}& 0.33 & 0.32 & 0.01& 0.16 &0.17\\
     \tt{off-1} & 0.34 & 0.11 & 0.23& 0.11 &0.23\\
     \tt{off-2} & 0.53 & 0.45 & 0.08& 0.22 &0.31 \\
     \tt{off-3} & 0.69 & 0.64 & 0.05& 0.30 & 0.39\\
     \tt{off-4} & 1.39 & 1.35 & 0.04& 0.23 & 1.16\\
    \bottomrule
    \end{tabular}
     \caption{\small A quantitative comparison of the L1 depth error [cm] when using keyframes and the full dataset for optimizing the network. We see that the keyframes are enough to learn the geometry of the scene and to generalize to unseen views.} \label{table:key-vs-full}
\end{table}

\subsection{Impact of camera pose errors on mapping}
In this section, we investigate the impact of ORB-SLAM pose errors on mapping quality. To achieve this, we first optimize the network using ORB-SLAM poses and then repeat the process with ground truth poses for the same number of iterations. We then evaluate the mapping quality of both networks on the same views. The results are summarized in \Cref{table:key-vs-full}. This experiment provides insight into the extent to which the observed error is attributable to camera poses versus the network's ability to accurately represent geometry. 

\subsection{Robustness}
In this experiment, we investigate the robustness of our online network optimization approach. Specifically, we optimize the network iteratively using the same ORB-SLAM camera poses multiple times. Our results summarized in \Cref{table:robust-eval} demonstrate that the network converges to the same results, indicating that our mapping approach is highly robust.

\begin{table}[ht]
  \centering%
\resizebox{0.48\textwidth}{!}{
  \begin{tabular}{llrrrrrr}
    \toprule
    & &  \multicolumn{1}{c}{\makecell{\tt{trial 1}}} & \multicolumn{1}{c}{\makecell{\tt{trial 2}}} &  \multicolumn{1}{c}{\makecell{\tt{trial 3}}} & \multicolumn{1}{c}{\makecell{\tt{trial 4}}} & \multicolumn{1}{c}{\makecell{\tt{trial 5}}} & Avg. \\
    \midrule    
    \multirow{4}{*}{\rotatebox[origin=c]{90}{room0}}
    & PSNR [dB]$\uparrow$ & 33.166 & 33.172 & 33.211 & 33.172 & 33.093 & 33.1630 $\pm$ 0.0381 \\
    & SSIM$\uparrow$ & 0.987 & 0.987 & 0.987 & 0.987 & 0.987 & 0.9870 $\pm$ 0.0000 \\
    & LPIPS$\downarrow$ & 0.016 & 0.016 & 0.016 & 0.016 & 0.017 & 0.0162 $\pm$ 0.0004  \\
    & L1 depth [cm]$\downarrow$ & 0.511 & 0.513 & 0.518 & 0.524 & 0.522 & 0.518 $\pm$ 0.0050 \\
    \midrule
    \multirow{4}{*}{\rotatebox[origin=c]{90}{room1}}
    & PSNR [dB]$\uparrow$ & 35.189 & 35.287 & 35.279 & 35.028 & 35.151 & 35.187 $\pm$ 0.0949 \\
    & SSIM$\uparrow$ & 0.988 & 0.988 & 0.988 & 0.987 & 0.988 & 0.9878 $\pm$ 0.0004 \\
    & LPIPS$\downarrow$ & 0.012 & 0.012 & 0.012 & 0.013 & 0.013 & 0.0124 $\pm$ 0.0004\\
    & L1 depth [cm]$\downarrow$ & 0.323 & 0.310 & 0.314 & 0.348 & 0.340 & 0.3270 $\pm$ 0.0147\\
    \midrule
    \multirow{4}{*}{\rotatebox[origin=c]{90}{room2}}
    & PSNR [dB]$\uparrow$ & 36.638 & 36.395 & 36.483 & 36.507 & 36.468 & 36.4982 $\pm$ 0.0793 \\
    & SSIM$\uparrow$ & 0.991 & 0.991 & 0.991 & 0.991 & 0.991 & 0.991 $\pm$ 0.0000\\
    & LPIPS$\downarrow$ & 0.011 & 0.011 & 0.011 & 0.011 & 0.011 &  0.011 $\pm$ 0.0000\\
    & L1 depth [cm]$\downarrow$ & 0.437 & 0.446 & 0.448 & 0.440 & 0.455 & 0.4452 $\pm$ 0.0063 \\
    \midrule
     \multirow{4}{*}{\rotatebox[origin=c]{90}{office0}}
    & PSNR [dB]$\uparrow$ & 40.120 & 40.259 & 40.225 & 40.110  & 40.432 & 40.2292 $\pm$ 0.1168\\
    & SSIM$\uparrow$ & 0.993 & 0.993 & 0.993 & 0.993 & 0.993 & 0.9930 $\pm$ 0.0000\\
    & LPIPS$\downarrow$ & 0.007 & 0.007 & 0.007 & 0.007 & 0.007 & 0.0070 $\pm$ 0.0000\\
    &L1 depth [cm]$\downarrow$ & 0.351 & 0.328 & 0.326 & 0.333 & 0.330 & 0.3336 $\pm$ 0.0090 \\
    \midrule
    \multirow{4}{*}{\rotatebox[origin=c]{90}{office1}}
    & PSNR [dB]$\uparrow$ & 38.865 & 38.847 & 39.106 & 38.870 & 38.853 & 38.9082 $\pm$ 0.0992\\
    & SSIM$\uparrow$ & 0.975 & 0.973 & 0.975 & 0.973 & 0.973 & 0.9738 $\pm$ 0.0009 \\
    & LPIPS$\downarrow$ & 0.018 & 0.019 & 0.017 & 0.020 & 0.020 & 0.0188 $\pm$ 0.0012 \\
    &L1 depth [cm]$\downarrow$ & 0.367 & 0.340 & 0.321 & 0.395 & 0.320 & 0.3486 $\pm$ 0.0288 \\
    \midrule
    \multirow{4}{*}{\rotatebox[origin=c]{90}{office2}}
    & PSNR [dB]$\uparrow$ & 34.201 & 34.224 & 34.188 & 34.319 & 34.216 & 34.2296 $\pm$ 0.0464\\
    & SSIM$\uparrow$ & 0.993 & 0.993 & 0.993 & 0.993 & 0.993 & 0.9330 $\pm$ 0.0000 \\
    & LPIPS$\downarrow$ & 0.009 & 0.009 & 0.009 & 0.008 & 0.009 & 0.0090 $\pm$ 0.0000\\
    & L1 depth [cm]$\downarrow$& 0.544 & 0.519 & 0.527 & 0.530 & 0.539 &  0.5318 $\pm$ 0.0090\\
    \midrule
    \multirow{4}{*}{\rotatebox[origin=c]{90}{office3}}
    & PSNR [dB]$\uparrow$ & 34.778 & 34.803 & 34.798 & 34.682 & 34.684 & 34.7490 $\pm$ 0.0545 \\
    & SSIM$\uparrow$ & 0.994 & 0.994 & 0.994 & 0.994 & 0.994 & 0.9940 $\pm$ 0.0000 \\
    & LPIPS$\downarrow$ & 0.006 & 0.006 & 0.006 & 0.007 & 0.007 & 0.0064 $\pm$ 0.0005 \\
    &L1 depth [cm]$\downarrow$ & 0.692 & 0.691 & 0.691 & 0.698 & 0.685 & 0.6914 $\pm$ 0.0041 \\
    \midrule
    \multirow{4}{*}{\rotatebox[origin=c]{90}{office4}}
    & PSNR [dB]$\uparrow$ & 33.263 & 33.270 & 33.233 & 33.268 & 33.178 & 33.2424 $\pm$ 0.0349 \\
    & SSIM$\uparrow$ & 0.989 & 0.988 & 0.989 & 0.989 & 0.989 &  0.9888 $\pm$ 0.0004\\
    & LPIPS$\downarrow$ & 0.015 & 0.015 & 0.015 & 0.015 & 0.015 & 0.0150 $\pm$ 0.0000 \\
    &L1 depth [cm]$\downarrow$ & 1.399 & 1.386 & 1.404 & 1.393 & 1.408 & 1.3980 $\pm$ 0.0078\\
    \bottomrule
  \end{tabular}
}
    \caption{\small Robustness evaluation on the Replica dataset. We see that our training is stable and the network reaches more or less the same quality for different runs.}\label{table:robust-eval}
\end{table}

\begin{table}[htp]
    \centering
    \begin{tabular}{lccc}
    \toprule
      & & \tt{L1 depth error(cm)} & \tt{PSNR(dB)} \\
    \midrule
    \multirow{2}{*}{room0} 
     & \tt{RGBD} & 0.51 & 33.16   \\
     & \tt{RGBD + sem.} & 0.50 & 32.16 \\
     \midrule
     \multirow{2}{*}{room1} 
     & \tt{RGBD} & 0.32 & 35.18   \\
     & \tt{RGBD + sem.} & 0.32 & 36.10 \\
     \midrule
     \multirow{2}{*}{room2} 
     & \tt{RGBD} & 0.44 & 36.49   \\
     & \tt{RGBD + sem.} & 0.44 & 36.06 \\
     \midrule
     \multirow{2}{*}{office0} 
     & \tt{RGBD} & 0.33 & 40.22   \\
     & \tt{RGBD + sem.} & 0.33 & 40.5 \\    
    \bottomrule
    \end{tabular}
    \caption{\small Effect of learning semantics on the accuracy of the mapping. We observe that adding the semantic branch does not decrease the quality of the learned geometry and color.}\label{table:semantics-vs-RGBD}
\end{table}

\begin{table}[H]
    \centering
    \begin{tabular}{cccccccccc}
    \toprule
      & \tt{rm-0} & \tt{rm-1}  & \tt{rm-2}  & \tt{off-0} \\
    \midrule
    NICE-SLAM \footnotesize{(no depth)} &11.12 & 9.42 & 19.03 & 11.12
    \\
    NeRF-SLAM \footnotesize{(RGB with noisy depth)} & 2.97& 2.63& 2.58& 2.49\\
     Ours \footnotesize{(RGBD)}& \textbf{0.46} & \textbf{0.29} & \textbf{0.43} & \textbf{0.31} \\
     Ours \footnotesize{(RGB)} & 3.3 &5.34 &3.5 &3.29 \\
     Ours \footnotesize{(RGB with semantics)} & 1.95 &  3.37 &  2.73 &  2.74 \\
    \bottomrule
    \end{tabular}
    \caption{\small Comparison between the L1 depth error [cm] of estimated depth quality of RGB and RGBD pipeline. The table shows that adding semantics can help to learn the underlying geometry more accurately. The scale ambiguity of RGB pipeline depth values is compensated based on the scale difference between RGB ORB-SLAM poses and ground truth poses. The value of other approaches are taken from \cite{nerf-slam}}\label{table:replica-depth-eval-seg}
\end{table}

\section{Conclusion}
This paper presents a novel SLAM algorithm for large indoor scenes, which combines classical tracking and loop-closing techniques with neural field-based mapping. Our approach results in accurate camera tracking and produces dense, memory-efficient maps of RGB, depth, semantics, and SDF data. We leave the extension of such SLAM algorithms to unbounded outdoor scenes as future work.

\clearpage
{\small
\bibliographystyle{IEEEtran}
\bibliography{semanticslam.bib}
}

 \end{document}